\documentclass[journal,10pt]{IEEEtran}
\usepackage{amsthm}
\usepackage{amsmath}
\usepackage{graphicx}
\usepackage{algorithm}
\usepackage{algorithmicx}
\usepackage{algpseudocode}
\usepackage{subfigure}
\usepackage{overpic}
\usepackage{colortbl}
\ifCLASSINFOpdf
\else
\fi
\begin{document}
\title{Multilevel Large Language Models for Everyone}

\author{Yuanhao~Gong\\College of Electronics and Information Engineering, Shenzhen University, China.~~gong.ai@qq.com% <-this % stops a space
	\thanks{Manuscript received April 19, 2005; revised September 17, 2014.}
	}


\markboth{Journal of \LaTeX\ Class Files,~Vol.~14, No.~8, August~2015}%
{Yuanhao: LLM on Blockchains}



\maketitle

\begin{abstract}
Large language models have made significant progress in the past few years. However, they are either generic {\it or} field specific, splitting the community into different groups. In this paper, we unify these large language models into a larger map, where the generic {\it and} specific models are linked together and can improve each other, based on the user personal input and information from the internet. The idea of linking several large language models together is inspired by the functionality of human brain. The specific regions on the brain cortex are specific for certain low level functionality. And these regions can jointly work together to achieve more complex high level functionality. Such behavior on human brain cortex sheds the light to design the multilevel large language models that contain global level, field level and user level models. The user level models run on local machines to achieve efficient response and protect the user's privacy. Such multilevel models reduce some redundancy and perform better than the single level models. The proposed multilevel idea can be applied in various applications, such as natural language processing, computer vision tasks, professional assistant, business and healthcare.
\end{abstract}


% \fi
\IEEEpeerreviewmaketitle

\section{Introduction}
%%%%%%%%% BODY TEXT
Large language models (LLM) are a fascinating result of recent advancements in artificial intelligence technology. These models have the potential to transform communication and writing. With the latest technological advancements, large language models have become more sophisticated in imitating human language. They are trained on enormous amounts of text data, enabling them to produce coherent and relevant text on various topics. This technology has opened up a wide range of usage in fields such as business, education, and healthcare.

In recent years, several popular large language models have emerged, each with unique qualities designed to cater to different requirements and perform different functions. For instance, GPT-3 generates highly accurate and convincing text, making it suitable for applications such as generating conversational responses or writing articles~\cite{Brown2020}. On the other hand, T5 is versatile and can perform a wide range of natural language processing tasks, making it valuable in areas such as translation and question answering~\cite{Raffel2020}.

Apart from GPT-3 and T5, BERT is another popular large language model that has made significant contributions to the field of natural language processing~\cite{Devlin2019}. BERT excels at understanding the context of words and sentences, making it valuable for applications such as search engines and chatbots. By understanding the context of sentences, BERT can provide more accurate and relevant search results, and it helps chatbots to produce more natural and human-like responses.

Despite their differences, all of these models share a common goal: to advance the field of artificial intelligence and push the limits of what machines can do. With the continued development of large language models, we can expect to see even more exciting breakthroughs in artificial intelligence technology in the future.

\begin{figure}
	\begin{minipage}{0.4\linewidth}
		\mbox{user level: personal LLM}
		\vfil
		\vspace{12mm}
		\mbox{field level: finance, IT}
		\vfil
		\vspace{4mm}
		\mbox{global level: all}
	\end{minipage}
		\begin{minipage}{0.6\linewidth}
	\centering
	\includegraphics[width=\linewidth]{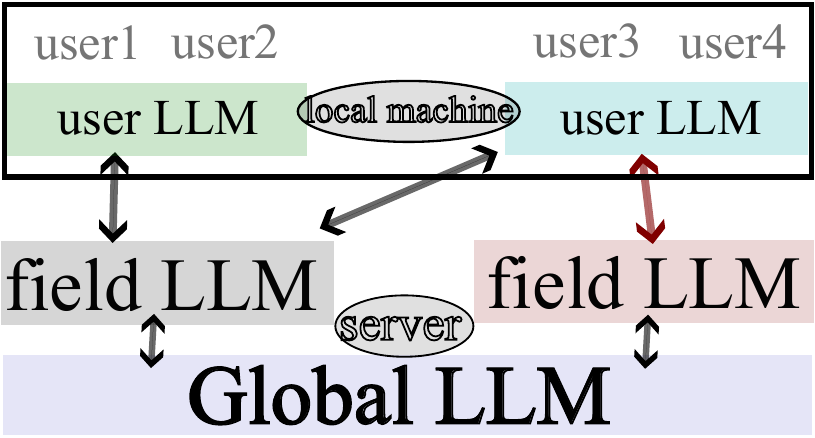}
\end{minipage}
\caption{Illustration of the proposed multilevel method. The global LLM are large and trained on a cloud. They are trained for the common intelligence. They are retrained (fine tuning) and distilled to get a smaller field specific LLM for different fields, such as medical diagnosis, mathematical education, and C++ programming. These models are further retrained and distilled to get a personal LLM that runs on a local machine to get efficient response and protect users' privacy from the internet attack.}
\label{fig}
\end{figure}

%Overall, while the trend towards larger and more complex neural networks presents challenges, there are also many opportunities for innovation and advancement in the field of artificial intelligence. The development of neural networks is an exciting field of research, and researchers are actively working to overcome the challenges posed by these models and unlock their full potential.
\subsection{Why Large Models?}
Large language models have transformed the field of natural language processing, allowing machines to understand and generate human-like language. These models learn from vast amounts of data and store and process more information due to their large number of parameters, enabling them to produce accurate, relevant, and contextually appropriate results. However, it is important to balance the number of parameters to prevent overfitting, which can reduce the model's overall performance. Additionally, training models with many parameters can be time-consuming, slowing down the training process.

Furthermore, large language models are versatile and can enhance various natural language processing tasks, such as language translation, sentiment analysis, and text summarization. By training on extensive data sets, these models can detect patterns and nuances in language that are difficult for humans to recognize. They are essential tools for applications such as virtual assistants, chatbots, and language understanding systems. Large language models can also generate human-like text, which has various applications in fields such as journalism, creative writing, and content generation. In conclusion, large language models are a powerful tool that can revolutionize how we interact with machines and generate human-like language, but it is important to balance the number of parameters and the model's performance for optimal results.
\subsection{Limitations}
Large language models have some limitations that need to be addressed. One such limitation is their potential to perpetuate biases present in their training data, which can lead to unfair and discriminatory outcomes. Additionally, these models have high computational and energy requirements, making them expensive to train and operate. Their tendency to generate unreliable or false information, particularly when dealing with complex or sensitive topics, is also problematic. Concerns have also been raised about the potential misuse of large language models for generating disinformation.

Another concern is the potential threat to privacy posed by large language models. Since these models require large amounts of data to train, they can potentially capture sensitive information. Therefore, it is important to develop privacy-preserving models that can ensure users' sensitive information is not compromised.

Furthermore, the environmental impact of running these models is also a concern. These models require significant amounts of energy to train and operate, leading to a significant carbon footprint. To mitigate this, researchers and developers are actively working to address these limitations and reduce the risks associated with large language models.

To address these limitations, researchers are developing more robust and diverse datasets for training. This can help reduce bias and enable models to produce more accurate results. Ethical guidelines are being implemented for model development and use to ensure that these models are used responsibly and do not cause harm. Finally, alternative approaches to language processing are being explored, such as using smaller models or leveraging other technologies like knowledge graphs to process language more efficiently.

%In conclusion, while large language models have some limitations, researchers and developers are working hard to address these issues and mitigate potential risks associated with these models.

\section{Multilevel Large Language Models}
The development of artificial intelligence has resulted in increasingly complex neural networks with more parameters. This has led to the creation of large language models such as GPT-3 and BERT, which have significantly impacted natural language processing.

However, a major challenge associated with these models is their high computational cost. Traditional large language models, like GPT-3 and BERT, require thousands of GPUs for training and deployment. This high cost is a significant obstacle for many researchers and organizations, limiting access to these powerful models.

Despite this challenge, advancements in hardware and software have made it possible to train and deploy these models on fewer GPUs. This has opened new opportunities for researchers and organizations to explore the capabilities of these models and leverage their potential in various applications.

Another problem with current LLM is that these models are static. In other words, they can not evolve to achieve higher performance with more and more user input. In fact, millions of users can help to improve the LLM in each specific field, such as poem, medical diagnosis and car design.

\subsection{Multilevel Models}
To tackle these issues, we propose a multilevel strategy. We propose to decompose the system into three levels: global level, field level and user level. Each level contains different LLM for different purpose. The full map is illustrated in Fig.~\ref{fig}. A simple comparison between these levels is shown in Table~\ref{table}.

At the global level, the LLM are trained using all available dataset. As a result, they are good at everything but not excellent at any specific fields. Letting users directly using these models is wasting the computation resource because the performance is just good, not excellent. However, such global LLM can help in improving the field LLM.

At the field level, the LLM are designed to be excellent in a specific field, such as poem, music, medical knowledge, and finance. These models correspond to the different professional career in human society such as lawyers and doctors. These models are excellent in a specific field in general, but not for any specific person.

At the user level, the LLM are trained to be an assistant with both personal information and professional advice from some specific field LLM. These models fill the gap between each individual user and the field LLM. Moreover, these models run on local machines to protect the user's privacy.  

\begin{table} 
	\centering \caption{Summary Multilevel LLM. }
	\begin{tabular}{c|c|c|c}
		\hline
		LLM & \# parameters & update period & inference time\\
		\hline
		global & large & long & long\\
		field & medium & medium & medium\\
		user & small & short & real-time\\
		\hline
	\end{tabular}
	\label{table}
\end{table}
\subsubsection{Global LLM}
The global LLM contain a very large parameters and are updated at low frequency. If nothing happens, they will automatically update themselves every week. However, if there are more information from the training data, they will update themselves with a shorter period.

Meanwhile, these global LLM are not designed to serve users. Instead, they are designed to update the field LLM that have specific domain knowledge and also smaller parameters. Therefore, users' response time is not affected by the long time training and inference processing from the global LLM.
\subsubsection{Field LLM}
The field LLM are designed to be professional in some field, such as art, music, object tracking, and education. They are automatically updated by the global LLM which receive new knowledge from the user input.

The field LLM can be considered as a buffer that exchanges information between the global LLM and the user LLM. Their parameter weights are updated by the global LLM. But they extract the information from the user LLM and feed the information into the global LLM. 

\subsubsection{User LLM}
At the user level, user LLM have much less parameters and also the inference time. Different from the global and field LLM, user LLM run on local machine and can contain each user's personal information such as movie and music preferences. Such user LLM can be considered as an excellent assistant who helps the user in planing, shopping, and working.

More importantly, these models are stored and encrypted on the user's local machine. Ideally, they will not betray the user. The user's privacy is protected.

\subsection{Dynamic Large Language Models}
Be aware that these models can evolve during users' usage. With more and more user input, the local LLM know more about the user's preference. Therefore, the local LLM can evolve to update themselves accordingly. They can also dynamically link themselves to several related field LLM because the user's input is also dynamically changing. Such dynamics affects the whole system, leading to dynamic LLM.  
\subsubsection{Learning during Using}
The multilevel large language models are designed to evolve during its usage. Thus, they are highly dynamic, especially with the user input. They are life-long learning systems, keeping evolving to understand the user from his/her history recording.
\subsubsection{Interaction between User and User LLM}
There is a direct interaction between users and users' LLM. The user input is dynamic and the user's preference might be also changing with time and experience. Such dynamics requires the user LLM quickly adapt to such change. 

To be adaptive, one way is to change the parameter or architecture in user LLM, based on the received users' input. The other way is to change the user LLM from the interaction with field LLM.
\subsubsection{Interaction between User LLM and Field LLM}
The user LLM and field LLM have a soft link. And such link can be changed dynamically when the user's preference is changed. For example, if the user's preference is changed from IT to finance, the user LLM will find such change and trigger a require to the field LLM. The field LLM will update the user LLM accordingly.

If the field LLM are updated, they will also update the user LLM to fresh the new knowledge into the user LLM. Such fresh frequency is medium to balance the model accuracy and the inference running time. 
\subsubsection{Interaction between Field LLM and Global LLM}
The field LLM can be updated by the user LLM and also the global LLM. Meanwhile, the global LLM can be updated by the field LLM that receive new information from the user LLM. The global LLM can be updated if there is new data set available.

Such interaction happens in a very low frequency because the global LLM are large and should be stable (not so adaptive to an individual user). In contrast, the user LLM are quickly adaptive to each user.
\section{Economical Ecosystem}
Our multilevel large language model (MLLM) is an innovative approach to training and deploying the LLM that offers significant benefits to both users and developers. By leveraging advanced computational techniques and sophisticated algorithms, the MLLM achieves unparalleled levels of accuracy and efficiency in natural language processing tasks. This translates to significant cost savings for developers, who can now train and deploy their models more quickly and easily than ever before. Meanwhile, users can enjoy a more seamless and intuitive experience when interacting with language-based applications and services, thanks to the MLLM's ability to understand and interpret natural language in a more nuanced and sophisticated way. With its cutting-edge technology and user-focused design, the MLLM is poised to revolutionize the field of natural language processing and usher in a new era of innovation and discovery.
\subsection{Users}
Users in the system have two roles. First, users are able to utilize the system's features, and as a result, they are required to pay a fee for the services rendered. The payment allows for continued access to the system and its benefits, including but not limited to increased productivity, streamlined processes, and enhanced data analysis capabilities. Additionally, regular payment ensures that the system can be maintained and further developed to meet the evolving needs of its users. Therefore, it is important to maintain a consistent payment schedule to ensure uninterrupted access to the system and its benefits.

Second, it is important to acknowledge the critical role that users play in the system. Users are valuable sources of input and integral to the system's success. Without their contributions, the system would lack quality and reliability. To encourage more high-quality contributors, incentivizing users to share their knowledge and expertise is an effective approach. One way to achieve this is by establishing a system of fair compensation for users, recognizing their efforts and contributions. This compensation can come in various forms, including monetary rewards, recognition, or exclusive access to features. By providing fair compensation, the system can attract a larger pool of motivated users who will contribute their best work, resulting in a more robust and reliable platform. Therefore, prioritizing the establishment of a fair compensation system for users is crucial, as it will ultimately benefit the system as a whole and contribute to its long-term success.
\subsection{Developers}
The developers can work on these three level LLM. They can develop the user LLM that can be more adaptive to each individual user. They can also develop field LLM, to make them more professional. They can work on global LLM to accelerate the model convergence and efficiency.

The developers also have two roles in this system. First, thanks to the extensive work on their LLM at these three levels, developers can earn money from their LLM. They are well-equipped to capitalize on their expertise and earn a lucrative income from their specialized model. With a deep understanding of the complexities of the field and the latest industry trends, they can confidently navigate the landscape and offer valuable services to a wide range of users. 

Second, they have to pay money for the (valuable) input from users. In order to obtain valuable input from their users, developers may pay a fee. This fee is necessary to ensure that the developer can continue to provide high-quality services and products that meet the needs and expectations of their customers. By compensating users for their input, developers can gain valuable insights that can help them improve existing products or develop new ones that better meet the needs of their target market. Additionally, this feedback can be used to identify areas for improvement in the developer's operations and customer service, which can lead to increased satisfaction and loyalty among existing customers and attract new ones.
\subsection{Dynamic Models}
The models in the system are not equal. Some are more important than others and thus have higher service fees. But this is also dynamic, which means that some important models might become important in the future. Therefore, all the price in the models are dynamic, according to their service quality and market demand. 

As a result, both users and developers have the potential to earn an income that is closely tied to the growth and progress of the system. As new technologies and innovations are introduced, users and developers have the opportunity to adapt and evolve LLM in order to stay competitive and continue to earn a sustainable income. This is different from the traditional economical mode for LLM, where the developers have to pay for the training and inference process first and then charge the users with their usage. 
\subsection{Blockchains}
From the hardware point of view, implementation of the proposed multilevel large language models is difficult. Each LLM requires a lot of GPUs to train. And there are many LLM in the multilevel large language models. The traditional way to train the system is unrealistic.

To tackle this issue, we propose to develop MLLM on blockchains, which have high computation performance, are decentralized and run parallel on all nodes. Interestingly, the decentralized mode in blockchains also fits the proposed local machine and server mode in the MLLM.

Each miner client in the blockchain can be considered as a user in the MLLM. Instead of only contributing the computation to the blockchain, the users also get services from the blockchain in the terms of computation and economy. Therefore the miner client and the blockchain charge each other according to their performance.

More importantly, the user's input is also considered as valuable resource that can improve the LLM at each level. Such economics is not considered in the traditional blockchains and traditional LLM. This novel idea will trigger the promotion of high quality input in the system and eventually benefit everyone in the ecosystem.

The system is still under developing and we will release the system in the future to accelerate the related research and practical applications. The economic mode behind MLLM can reduce the barrier for LLM development, making the developing easier and available for everyone. 

Nowadays, the mobile phones and chips become more and more powerful. They are likely the node in the blockchains. And both the users and developers should benefit from such devices and get higher quality services.
\section{Conclusion and Discussion}
In recent times, there has been a significant rise in the usage of large language models. These models are created to comprehend and analyze human language and are becoming increasingly popular in various industries such as natural language processing, sentiment analysis, machine translation, and speech recognition. With the increasing demand for these models, we can expect to see further developments in the field of artificial intelligence and natural language processing. These advancements will undoubtedly impact the way we interact with technology and communicate in the future.

Although the large language models have made significant progress in natural language processing, there are still some limitations that need to be addressed. One of the most important limitations is the computational power needed to train and run these models. In addition, while these models can generate coherent text, they often struggle with generating text that is both relevant and accurate. Another limitation is the lack of diversity in the training data used to create these models, which can lead to bias and inaccuracies in the generated text. Therefore, it is essential to continue exploring ways to improve these models, such as incorporating more diverse and representative training data and developing more powerful computational resources to train and run these models.

In this paper, we have introduced multilevel large language models that address some of the challenges posed by previous large language models. Specifically, our models incorporate innovative strategies for handling long-term dependencies, improving computational efficiency, and enhancing the accuracy of predictions. Overall, our work represents a significant step forward in the development of large language models and lays the groundwork for future research in this area.

The multilevel approach seeks to balance accuracy and running time. It is based on the concept that complex problems can be broken down into smaller, simpler sub-problems to achieve accurate results. By solving these sub-problems, the overall accuracy of the solution can be improved. However, creating too many sub-problems can increase the running time. The multilevel approach suggests that a balance can be struck by dividing the problem into an optimal number of sub-problems. This allows for the maintenance of accuracy while keeping the running time within acceptable limits. The multilevel approach has gained popularity in various fields, such as computer science, mathematics, and engineering. It has been successfully applied to problems such as image processing, signal analysis, and optimization. Overall, the multilevel approach is a promising strategy that can improve the accuracy of complex problem solutions while keeping the running time practical.
%Finally, blockchain can be used to support the decision-making process in AI systems. By creating a decentralized network of nodes that vote on decisions, we can ensure that decisions are made in a fair and transparent manner.

Our approach is a novel step forward in the development of next generation large language models. The multilevel methods also show a novel economic model for developing large language models. They can be adopted in various applications such as audio, vision and machine learning tasks \cite{chenouard:2014,gong2009symmetry,Lewis2019,zhao2023survey,Gong2012,Brown2020,gong2013a,Yu2019,Gong:2014a,Yin2019a,gong:phd,Yu2022a,gong:gdp,Guo2022,gong:cf,Zong2021,gong:Bernstein,Ezawa2023,Gong2017a,Tang2021a,Gong2018,Gong2018a,Yu2020,GONG2019329,Sancheti2022,Gong2019a,Tang2021,Gong2019,Yin2019b,Gong2022,Yin2020,Gong2020a,Jin2022,Gong2021,Tang2022,Gong2021a,Tang2022a,Tang2023,Gong2022,Tang2023a,Xu2023,Han2022,Scheurer2023,Zhang2023b}.
\bibliographystyle{IEEEtran}
\bibliography{IEEEabrv,../IP}



%\vfill


\end{document}